\documentclass[journal]{IEEEtran}

\usepackage{hyperref}
\usepackage{cite}

\usepackage{graphicx}

\usepackage{algorithm}
\usepackage[noend]{algpseudocode}
\usepackage{cite}

\usepackage{amssymb}
\usepackage{amsmath}

\usepackage{dsfont}

\usepackage{comment}
\usepackage[shortlabels]{enumitem}

\usepackage{threeparttable}

\usepackage[caption=false,font=footnotesize]{subfig}

\usepackage{fixltx2e}
\usepackage{stfloats}
\usepackage{siunitx}

\usepackage{multirow}

\newtheorem{theorem}{Theorem}

\newtheorem{definition}{Definition}

\newcommand{\argmin}{\mathop{\rm argmin}}

\usepackage{xcolor}
\newcommand{\Gao}[1]{{\color{black} #1}}

\newcommand{\KH}[1]{{\color{black} #1}}
\newcommand{\Mod}[1]{{\color{black} #1}}

\usepackage{arydshln}
\makeatletter
  \renewcommand*\env@matrix[1][*\c@MaxMatrixCols c]{%
    \hskip -\arraycolsep
    \let\@ifnextchar\new@ifnextchar
  \array{#1}}
\makeatother

\graphicspath{{Images/}}
\begin{document}
%
\title{ Fast UAV Trajectory Optimization using Bilevel Optimization with Analytical Gradients }

\author{Weidong~Sun\textsuperscript{*},
        Gao~Tang\textsuperscript{*},~\IEEEmembership{Student Member,~IEEE},
        and~Kris~Hauser,~\IEEEmembership{Senior Member,~IEEE}
\thanks{W. Sun is with XYZ Robotics Inc. Shanghai 200000 China. e-mail: weidong.sun@xyzrobotics.ai}
\thanks{G. Tang and K. Hauser are with the Departments of Computer Science, University of Illinois at Urbana-Champaign. Urbana, IL, 61820 USA. e-mail: {gaotang2, kkhauser}@illinois.edu.}
\thanks{\textsuperscript{*} Denotes equal contribution }
\thanks{This work is partially supported by NSF Grant \#IIS-1253553. W. Sun is partially supported by the China Scholarship Council (CSC).  This work was conducted in part while the authors were affiliated with Duke University. }
}

\maketitle

\begin{abstract}
We present an efficient optimization framework that solves trajectory optimization problems by decoupling state variables from timing variables, thereby decomposing a challenging nonlinear programming (NLP) problem into two easier subproblems. With timing fixed, the state variables can be optimized efficiently using convex optimization, and the timing variables can be optimized in a separate NLP, which forms a bilevel optimization problem. The challenge of obtaining the gradient of the timing variables is solved by sensitivity analysis of parametric NLPs. The exact analytic gradient is computed from the dual solution as a by-product, whereas existing finite-difference techniques require additional optimization. The bilevel optimization framework efficiently optimizes both timing and state variables which is demonstrated on generating trajectories for an unmanned aerial vehicle. Numerical experiments demonstrate that bilevel optimization converges significantly more reliably than a standard NLP solver, and analytical gradients outperform finite differences in terms of computation speed and accuracy. Physical experiments demonstrate its real-time applicability for reactive target tracking tasks.
\end{abstract}

\section{Introduction}

\IEEEPARstart{R}{eal-time} optimal trajectory generation has long been a challenging but essential component in robotics. Due to the prevalence of nonlinear dynamics, non-convex constraints and high dimensionality of the planning space, it is often difficult to optimize trajectories quickly and reliably. One promising approach is to fix a subset of optimization variables and optimize the rest by convex optimization that can be reliably solved to global optimum. One such example is to fix the time parametrization of a polynomial spline trajectory, which permits optimizing the trajectory with quadratic programming (QP) methods. This technique has been applied to path planning for ground robots, autonomous cars~\cite{wang2018safe,fan2018baidu},  humanoid robots~\cite{fernbach2018croc} and UAVs~\cite{mellinger2011minimum,gao2018online,Liu2017PlanningDF}.
However, this method's optimality relies on the choice of a proper time parametrization of the splines, which, given its intricate and highly nonlinear relationship with the optimization objective and the problem's constraints is not trivial. Indeed, most existing methods rely on heuristics for that task \cite{gao2018online} and efficient optimization of time-allocation still remains an open question despite efforts in \cite{mellinger2011minimum,richter2016polynomial}.
The lack of ability to efficiently optimize time allocation leads to two situations. 
In one situation the optimization takes a long time and is thus unable to be used in highly reactive problems.
In the other one, heuristics are used and the trajectory has large jerks, which is more likely to fail due to thrust limit and consumes more energy.
It is thus necessary to efficiently optimize the time allocation of UAV trajectories.

\begin{figure}[t!]
  \centering
  \subfloat[Trajectories flying through a point cloud environment with the initial trajectory (red) and optimized trajectory using our method (black). The 12 boxes indicate the safe corridor in which the UAV is constrained.]{\includegraphics[width=0.4\textwidth]{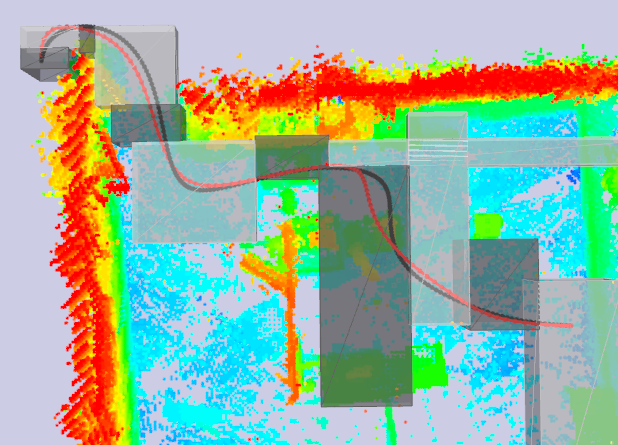}
  \label{fig:gazebo_example:pointcloud}}
  \vfill
  \subfloat[Velocity profiles of the trajectories in along the $x$ and $y$ direction.]{\includegraphics[width=0.45\textwidth]{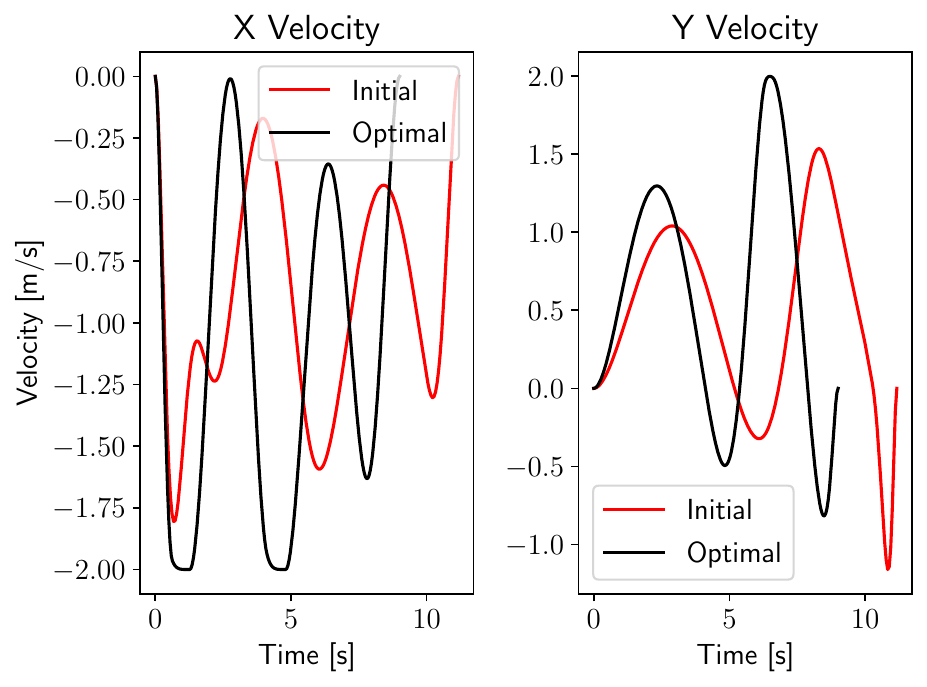}
  \label{fig:gazebo_example:acc}}
  \caption{ Initial trajectory is computed by the heuristic time assignment from Gao et al.~\cite{gao2018online}. Our method optimizes a weighted sum of total jerk and time. }
  \label{fig:gazebo_exapmle}
  \vspace{-0.2cm}
\end{figure}

One approach to this probem was proposed by Mellinger et al.~\cite{mellinger2011minimum}, who applied a gradient descent method to refine time allocation. Specifically, they divided the  optimization problem in two subproblems (or levels): the lower level optimizes the path while timing is fixed using QP, and the upper level optimizes the time allocation using gradient descent. Similar approach is used in \cite{richter2016polynomial}.

Nonetheless, finding the gradient of the cost function w.r.t. the time allocation \Gao{with constraints present in the QP} remains an unresolved issue \Gao{even though analytic gradient is recently given for unconstrained QP \cite{wang2020generating}.} To address this, finite difference method \cite{mellinger2011minimum} has been employed. But it can be computationally expensive since the number of QPs that needs to be solved at each gradient step grows linearly with the number of spline segments. Moreover, the gradient is inaccurate due to truncation errors and difficulty in the choice of step size.
As a result, using finite difference takes longer to converge and tends to converge to a worse cost.

Aiming to address this complexity, we use sensitivity analysis techniques to compute the gradient of the time allocation from the dual solution (Lagrange multipliers) of the QP. 
By exploiting the dual solution of the QP, our method allows us to compute exact analytical gradients w.r.t. time allocation, bypassing the downsides of finite difference methods: high computational complexity and low accuracy.
The gradient of time allocation is used with gradient descent method to optimize the time allocation.

Our framework combines bilevel optimization and analytic gradient.
It decouples spatial and temporal variables and solves them hierarchically. 
The lower level solves the spatial variables and utilizes off-the-shelf convex solvers to handle the strong sensitivity of polynomial coefficients and constraints for collision avoidance.
The upper level optimizes temporal variables that are numerically better behaved with simple linear constraints using gradient descent with line search and projection.

One potential issue is the non-smoothness of the upper level optimization. Fiacco \cite{fiacco1983introduction} shows that the smoothness of the optimal cost of parametric optimization problems, i.e. objective function of the upper level optimization, requires smoothness of both the cost and constraints, strong convexity near optimum, and some constraint qualifications.
In fact, these constraint qualifications do not hold universally and as a result the upper level is theoretically a non-smooth optimization problem.
Although this is not a problem in practice, for completeness our method treats objective function discontinuities using a subgradient method when non-differentiability is detected preventing the progress of gradient descent.
We prove the convergence of this method despite potential function non-smoothness.

The bilevel optimization framework is guaranteed to yield feasible results at any point in its execution, allowing for arbitrary termination. In contrast, alternative solution methods based on nonlinear optimization such as direct collocation and joint optimization of spatial and temporal variables have no such guarantee due to the problem's strong nonlinearity. Even with feasible initial guesses, our experiments show that these methods do not have a high success rate.
Compared with straightforward gradient approximation by finite difference, our analytic gradient is more accurate and computationally efficient.
As a result, our method outperforms finite-difference baselines in terms of both computation speed and solution quality.
One numerical example showing the effectiveness of our technique is in Fig. \ref{fig:gazebo_exapmle}. 
Physical experiments using a real quadrotor on point-to-point navigation and dynamic goal tracking further demonstrate the effectiveness and real-time capabilities of our method.

This paper is an extension of our conference publication \cite{accversion} where the contributions include
\begin{enumerate}
    \item A bilevel optimization framework to optimize time allocation with analytic gradient.
    \item Comparison with finite-difference, direct collocation, and joint optimization.
\end{enumerate}
In this paper the new contributions include
\begin{enumerate}
    \item Demonstration of real-time capability in physical experiments.
    \item Theoretical justification of the theorem being used.
    \item Subgradient descent method to handle non-smoothness of the cost function and proof of convergence.
    \item Test of the algorithm in more complex environments and realistic environments represented by point clouds.
    \item Theoretical and empirical scalability analysis to problems with more than 40 segments.
\end{enumerate}

\section{Related Work}
\subsection{Trajectory Optimization for UAVs}
Trajectory optimization solves the problem of computing the optimal trajectory for dynamic systems under some cost function and constraints.
It is a widely used method for motion planning of robotic systems. 
See \cite{betts1998survey} for an introduction of trajectory optimization for general systems.
Specifically, direct collocation \cite{von1993numerical} has been widely used.
However, pioneered by Mellinger \cite{mellinger2011minimum}, polynomial trajectories which exploit the differential flatness of UAV dynamics are often used for UAVs.
With the snap of the trajectory as the cost function, a quadratic optimization suffices to compute the optimal trajectory.
Some extensions to this framework include using B\'ezier curve control points as optimization variables and safe corridor generation to guarantee collision avoidance \cite{gao2018online}.
However, to use this framework, the temporal variables of the trajectory such as the duration of each segment have to be chosen prior to optimization.
They are usually chosen by heuristics and this provides room for optimization of time allocations.
Wang et al.~\cite{wang2020alternating} propose an alternating method but it lacks the ability to handle complex spatial constraints. 
\Gao{
Recent work \cite{tordesillas2019faster} uses mixed-integer QP to solve for trajectories and guarantees safety by always having a feasible,
safe back-up trajectory. 
However, the mixed-integer QP does not scale well with the number of convex polyhedra.
Our paper demonstrates the scalability of our method, which can solve problems in corridors with up to more than 40 polyhedra.
Gao et al.~\cite{gao2020teach} computes safe and aggressive trajectories in real time from human-piloted trajectories, and generates them by alternatively optimizing the spatial and temporal trajectories. 
However, the two parts optimize different objective functions, so convergence is not guaranteed.
On the contrary, our bilevel optimization method has been proven to converge to a local minimum of a unified objective.
}

\subsection{Optimizing Time Allocation}
As described above, trajectory optimization with polynomial splines is a well-studied problem as long as the spline timing is fixed. However, finding an optimal time allocation in real-time is still challenging. One strategy~\cite{gao2018online, Liu2017PlanningDF} is to use heuristics such as graph search on a discretized grid to generate a time allocation and keep timing fixed during the optimization stage. 
Heuristics in general are not optimal and can lead to inefficient trajectories, as shown in Fig. \ref{fig:gazebo_exapmle} where the initial trajectory has larger jerks than the optimum. 
Iterative methods such as gradient descent \cite{mellinger2011minimum,richter2016polynomial} have also been used to optimize time allocation.
However, if gradients are computed by finite differences, $(n+1)$ QPs have to be solved for a problem with $n$ segments for every gradient evaluation, making it slow and inaccurate.
Another strategy to determine time allocation is to use sampling~\cite{fernbach2018croc}. This approach randomly samples the duration of each spline segment until the corresponding QP can be solved, and has been applied successfully to humanoid locomotion problems.  Because optimality is not emphasized, this method is only helpful in settings where obtaining a feasible solution is the major bottleneck.

\subsection{Bilevel Optimization}

\text{Bilevel optimization}~\cite{sinha2018review} refers to a mathematical program where one optimization problem (the upper-level optimization problem) has another optimization problem (the lower-level optimization problem) as one of its constraints, i.e., one optimization task is embedded within another. 

Bilevel and multi-level optimization techniques have been employed for switching time optimization for switched systems~\cite{xu2004optimal, egerstedt2003optimal, johnson2011second, farshidian2017efficient}. These works focus on calculating derivatives of an objective function with respect to switching times.
In particular, works by Xu et al.~\cite{xu2004optimal} and Egerstedt et al.~\cite{egerstedt2003optimal} compute the derivatives using Lagrange multiplier methods, which bear some resemblance to sensitivity analysis technique used in this paper. However, these works are based on Pontryagin's Maximum Principle~\cite{betts1998survey} and fall short of the capability to include inequality path constraints, which is often unavoidable in robotic applications.

Applications of bilevel optimization in robotics include trajectory optimization for legged robots~\cite{farshidian2017efficient}, and robust control and parameter estimation~\cite{landry2019differentiable}. Landry et al.~\cite{landry2019differentiable} present a bilevel optimization solver based on an augmented Lagrangian method, however their bilevel method is slower than directly solving the NLP in their experiments.

Many algorithms are available to solve bilevel optimization, and we refer readers to Sinha et al.~\cite{sinha2018review} and Colson et al.~\cite{colson2007overview} for more comprehensive treatments of the topic. Most closely related to our approach is the \textit{descent method}, which seeks to decrease the upper-level objective while keeping the new point feasible. Our method might be categorized as a descent method as we solve the upper-level optimization problem by gradient descent using gradients provided by the lower-level optimization problem. 

\subsection{Gradients in Bilevel Optimization}
Efficiently and accurately computing gradients of the lower-level optimization problem is essential in applying gradient-based methods to solve bilevel optimization problems. The key derivation used in our algorithm is based on sensitivity analysis for parametric NLPs~\cite{fiacco1983introduction}. The method used to compute gradient in this work is very similar to Pirnay et al.~\cite{pirnay2012optimal}, which provides the optimal sensitivity of solutions to NLP problems. 

Efficient computation of gradients of optimization problems is also explored in the field of machine learning. OptNet~\cite{amos2017optnet} incorporates a QP solver as a layer into the neural network and is able to provide analytical gradients of the solution to the QP with respect to input parameters for back propagation. 
Gould et al.~\cite{gould2016differentiating} presents results on differentiating argmin optimization problems with respect to optimization variables in the context of bilevel optimization.

\section{Methodology}

In this section, we describe our framework. We start by a mathematical formulation of the trajectory optimization problem for UAVs and end with a description of our algorithm.

\Mod{
\subsection{Trajectory Optimization Preliminaries}

In the case of UAV motion planning, differential flatness allows us to plan a trajectory in the UAV's four flat outputs $[x(t)^T \; \psi(t)]^T$ which consist of 3-D position $x(t) \in \mathbb{R}^3$ and yaw angle $\psi(t) \in SO(2)$, without explicitly enforcing dynamics~\cite{mellinger2011minimum}. In this work, we plan in $\mathbb{R}^3$ by assuming the yaw stays constant, which is a common practice in UAV motion planning.

} 

Trajectory optimization aims to find a trajectory $x:[0,T] \rightarrow \mathbb{R}^d$ that minimizes some measure of performance $J$ while satisfying all the necessary constraints, e.g. being collision-free and dynamically feasible, and is formulated as:

\begin{equation*}
    \begin{array}{ll}
    \underset{x, T}{\mbox{minimize}}  & J(x,T) = {\displaystyle \int_0^T} \ell(x,t) dt + \Phi(x(T)) \\
    \mbox{subject to} & x(0) = x_0 \\
    & x(T) \in \mathcal{X}_\textrm{goal} \\
    & g(x(t)) \leq 0, \quad \forall t\in[0,T] \\
    & h(x(t)) = 0,  \quad \forall t\in[0,T], \\
    \end{array}
    \label{problem:trajectory_optimization}
\end{equation*}

where $x$ encodes the trajectory, with the initial state $x_0$ and goal region $\mathcal{X}_\textrm{goal}$ prescribed.  Total traversal time is denoted as $T$, which is sometimes fixed.  The objective $J(x,T)$ is the sum of running cost ${\displaystyle \int_0^T} \ell(x,t) dt$ and terminal cost $\Phi(x(T))$. Constraints including dynamics, collision avoidance, and other system constraints are encoded in $g(\cdot)$ and $h(\cdot)$.  

\Mod{
\subsection{Safe Corridors}
\label{subsection:corridor}

To guarantee that the whole trajectory will stay collision-free, we extract a safe corridor from the environment using the implementation from Gao et al.~\cite{gao2018online}. Here we give a brief overview of their method. Given a map represented by an occupancy grid or an Euclidean signed distance field (ESDF), a start position and a goal position, a safe corridor is generated by taking the following steps:
\begin{enumerate}
    \item Inflate all the obstacles in the map by a safety radius, so that the UAV can be considered as a point.
    \item Find a path that connects the start and the goal using $A^\star$ search or fast marching method (FMM).
    \item Grow a safe corridor consisting of convex polytopes around the path. In our implementation, we generate axis-aligned boxes by growing an axis-aligned box centered around each node in the path until it hits an obstacle, and then removing redundant boxes.
\end{enumerate}

One such corridor and its corresponding environment \footnote{\url{http://ais.informatik.uni-freiburg.de/projects/datasets/octomap/}. Last retrived Jul-31-2020.} is shown in Fig.~\ref{fig:gazebo_example:pointcloud}.
Here, axis-aligned boxes are chosen for simplicity and compatibility with the grid data structures commonly used by perception algorithms. But in general, our method is applicable to corridors composed of any convex polytope. } 

\subsection{Trajectory Optimization with Piecewise B\'ezier Curves}
\Mod{
In this section we define a trajectory optimization problem in terms of spatial variables $c$ (polynomial coefficients) and temporal variables $y$ (durations of each segment of the curve). 

We represent the trajectory as a piecewise B\'ezier curve of order $d$ with $n$ segments and segment durations $\Delta t_1,\ldots,\Delta t_n$. The timing of each knot point (connection point between two consecutive pieces) is given by $t_i = t_{i-1} + \Delta t_i$ with $t_0=0$.  The $i$'th segment is defined over the domain $[t_{i-1},t_i]$ as:
\begin{equation*}
    x(t) = \sum_{j=0}^d c_{ij} B_{d,j}\left(\frac{t-t_{i-1}}{\Delta t_i}\right), \quad t\in [t_{i-1}, t_i]
\end{equation*}
for each $i=1,\ldots,n$, where $c_{ij} \in \mathbb{R}^3$ denotes the $j$'th control point in the $i$'th segment and $B_{d,j}$ denotes the $j$'th Bernstein polynomial of order $d$ defined as
\[
B_{d,j}(u) = \dfrac{d!}{j! (d-j)!} u^j (1-u)^{d-j}.
\]

We gather all the polynomial coefficients (control points) in the flattened vector $c \in \mathbb{R}^{3n(d+1)}$ and define the time allocation as $y = [\Delta t_1, \ldots, \Delta t_n]^T \in \mathbb{R}_{+}^n$ .
} 

\subsubsection{Objective Function}

Often, the objective function $J$ is chosen to be the integral of the squared norm of some high-order derivative of the trajectory to penalize control effort. We use a more general objective:
\begin{equation}
    J(x,T) =  \int_0^T \|x^{(q)}(t) \|^2 dt + w T, 
    \label{eq:objective_first}
\end{equation}
which is a weighted sum of the integral of the squared $L_2$-norm of the $q$'th derivative and the traversal time $T$, with weighting parameter denoted by $w$. 

It has been shown that the first term in Eq.~\eqref{eq:objective_first} can be written as a quadratic function of the coefficients $c$, with the quadratic matrix $P_q(y)$ determined by time allocation $y$ and the order of derivative $q$ ~\cite{mellinger2011minimum,gao2018online}. 
With a slight abuse of notation, we can write Eq.\eqref{eq:objective_first} in terms of polynomial coefficients $c$ and time allocation $y$:
\begin{equation}
    J(c,y) =  c^T P_q(y) c + \mathds{1}^T y,
    \label{eq:objective}
\end{equation}
where $P_q(y)$ is a symmetric positive semidefinite matrix that is nonlinear in $y$ and $\mathds{1}$ is a vector of 1s. Note that we will drop the subscript $q$ in $P_q(y)$ from now on since it is assumed to be $q=3$ (we wish to find a minimum jerk trajectory) throughout this work.

\subsubsection{Constraints on Continuity}

Constraints on the trajectory should be enforced so that:
\begin{enumerate}[a)]
    \item States at the start and end of the trajectory should match the initial state and (optional) final state.
    \item Continuities at knot points which ensure a smooth transition between each segment of the trajectory. If the trajectory needs to be $C^{k}$ continuous, equality constraints up to the $k$'th order should be applied at all knot points. We found that in the UAV case, applying continuity constraints up to acceleration yields good results, the same as \cite{gao2018online}.
\end{enumerate}
The above constraints can be compiled into a linear equality constraint on the polynomial coefficients $c$:
\begin{equation}
    H(y) c = m,
    \label{eq:Lc_eq_m}
\end{equation}
where the matrix $H$ is generally nonlinear in time allocation $y$.

\subsubsection{Constraints on Safety and Dynamic Feasibility}
Safety and dynamical feasibility are ensured by imposing inequality constraints such that:
\begin{enumerate}
    \item The whole trajectory stays in the safe corridor discussed in Section.~\ref{subsection:corridor}.
    \item The maximum velocity $\| x^\prime (t) \|_\infty$ and maximum acceleration $\| x^{\prime \prime} (t) \|_\infty$ along the trajectory are bounded, i.e.,
    \begin{equation}
        \| x^\prime (t) \|_\infty \leq v_\textrm{max}, \quad \| x^{\prime \prime} (t) \|_\infty \leq a_\textrm{max} \quad \forall t \in [0, T]
        \label{eq:dynamic_feasibility}
    \end{equation}
    with $v_\textrm{max}$ and $a_\textrm{max}$ prescribed by the capabilities of the vehicle, user preference or operational norms.
\end{enumerate}

We encode the trajectory using a piecewise B\'ezier curve~\cite{gao2018online}, which has the properties:
\begin{enumerate}
    \item The curve is totally contained in the convex hull of its control points.
    \item The derivative of a B\'ezier curve is again a B\'ezier curve, with its coefficients being a linear combination of its antiderivative's coefficients.
\end{enumerate}
Using these properties safety and dynamically feasible constraints can be imposed as a linear inequality constraint on the flattened coefficients  ~\cite{gao2018online}:
\begin{equation}
    G(y) c \leq h,
    \label{eq:Gx_leq_h}
\end{equation}
where matrix $G$ is generally nonlinear in time allocation $y$.

We note that constraining position, velocity and acceleration using control points of B\'ezier curve does introduce some conservativeness since only the ends of the curve reach the convex boundary even if all control points are at boundary. \Gao{This effect can be seen in Fig.~\ref{fig:gazebo_example:acc}, in which the  velocity limits of $\pm$2\,m/s are only reached at a few discrete points.}
\Gao{
An alternative collocation implementation would use a grid along the trajectory, with constraints applied at grid points.  Although collocation is less conservative, it would not guarantee feasibility at non-grid points. 
Another approach is to split the B\'ezier curve into more pieces so the curve can be represented by more control points.
We refer readers to \cite{tordesillas2020minvo} for a more detailed discussion about the conservativeness of B\'ezier curve in a convex hull and potential alternatives. 
}
If the total trajectory time can be changed, one can optimize without velocity and acceleration bounds and simply increase the total trajectory time to satisfy the bounds without losing optimality of the time allocation. 
In fact, since our objective minimizes jerk, the dynamic feasibility is already considered in the cost function to some extent. 
No matter what representation is used, only the lower level optimization is affected and the efficacy of our analytic gradient computation and gradient descent method still holds. 

\subsubsection{Constraints on Time}
Hard constraints on time allocation $y$ may be imposed, such as a fixed total traversal time or that the duration of each segment must be positive. We encode these constraints as
\begin{equation}
    A y \leq b, \quad C y = d,
    \label{eq:Ay_leq_b}
\end{equation}
with $A, b, C, d$ properly chosen.

\subsection{Final Formulation}
In summary, we collect Eq.~\eqref{eq:objective}, \eqref{eq:Lc_eq_m}, \eqref{eq:Gx_leq_h} and \eqref{eq:Ay_leq_b}, \Mod{into the problem of Trajectory Optimization using B\'ezier spline in a Corridor (TOBC):}
\begin{equation*}
    \begin{array}{ll}
    \underset{c, y}{\mbox{minimize}}  & J(c,y) = c^T P(y) c + w \mathds{1}^T y \\
    \mbox{subject to} & Ay \leq b \\
    & C y = d \\
    & G(y) c \leq h \\
    & H(y) c = m,
    \end{array}
    \tag{TOBC}
    \label{problem:bezier_optimization}
\end{equation*}
which is nonlinear in time allocation $y$ and convex (quadratic) in spline coefficients $c$ for fixed $y$. This formulation generalizes the formulations found in ~\cite{fan2018baidu,fernbach2018croc,gao2018online,Liu2017PlanningDF}. 

We mainly study two variants of \eqref{problem:bezier_optimization}.  The Hard Time variant imposes a fixed traversal time and uses minimum-jerk as the objective (following Mellinger and Kumar~\cite{mellinger2011minimum}):
\begin{equation}
    \begin{array}{lll}
    \underset{c, y}{\mbox{minimize}}   & J(c,y) = c^T P(y) c \\
    & \mathds{1}^T y = T_c \\
    & y \geq \delta \\
    & G(y) c \leq h \\
    & L(y) c = m,
    \end{array}
    \label{problem:bilevel_optimization:min_jerk}
\end{equation}
where $T_c$ is a fixed traversal time, e.g., chosen by a higher-level planner, and $\delta$ is a small value (we use \num{1e-6}) which ensures that durations are positive in each corridor. We note that the fixed traversal time $T_c$ is necessary.

The Soft Time variant uses a weighted sum of jerk and traversal time as the objective (following Richter et al.~\cite{richter2016polynomial})  rather than imposing hard constraints on traversal time:
\begin{equation}
    \begin{array}{lll}
    \underset{c, y}{\mbox{minimize}}   & J(c,y) = c^T P(y) c + w \mathds{1}^T y \\
    & y \geq \delta \\
    & G(y) c \leq h \\
    & L(y) c = m.
    \end{array}
    \label{problem:bilevel_optimization:time_penalty}
\end{equation}

One use case of Eq.~\eqref{problem:bilevel_optimization:time_penalty} is tracking a dynamic goal, with the tracking aggressiveness tuned by the weight $w$.

\subsection{Formulation of the Bilevel Optimization Problem}

\begin{figure}[!t]
\centering
\includegraphics[width=3.0in]{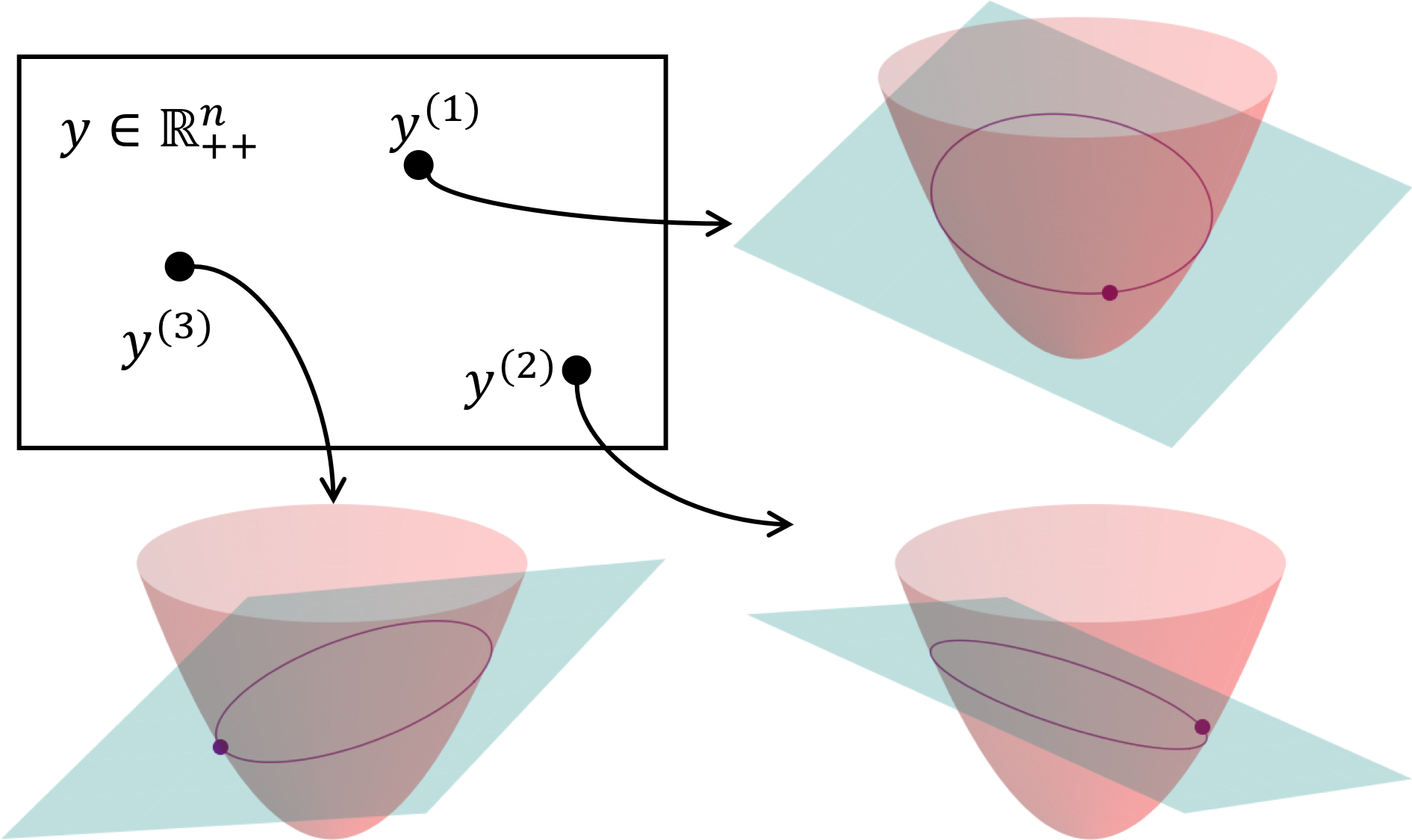}
\caption{An illustration of bilevel optimization: $y^{(1)}$, $y^{(2)}$ and $y^{(3)}$ are three feasible time allocations, they are optimized in the upper-level optimization problem. Each of these $y$ corresponds to a quadratic programming problem, which is being solved in the lower-level optimization problem. The red paraboloids are quadratic objective functions, cyan planes are equality constraints, no inequality constraints are drawn for illustration purposes. Feasible sets are purple curves, and purple dots are the optimal solutions to each lower-level optimization problem.}
\label{fig:bo}
\end{figure}

To efficiently solve \eqref{problem:bezier_optimization}, we will rewrite it as a bilevel optimization problem, which is defined as follows \cite{sinha2018review}.
\begin{definition}
\label{def: definiton of a bilevel optimization problem}
A bilevel optimization problem is given by
\begin{equation*}
    \begin{array}{cll}
    \underset{x_u \in X_U, x_l \in X_L}{\mbox{minimize}}   & F(x_u, x_l) \\
    \mbox{subject to} & x_l \in \underset{x_l \in X_L}{\argmin} \{ &f_0(x_u, x_l): \\ &&f_i(x_u, x_l) \leq 0,i=1,\ldots, m \\
    &&h_i(x_u, x_l) = 0, i=1, \dots, p\} \\
    &\multicolumn{2}{l}{G_i(x_u, x_l) \leq 0, \quad i=1, \ldots, M} \\
    &\multicolumn{2}{l}{H_i(x_u, x_l) = 0, \quad i=1, \ldots, P}
    \end{array}
\end{equation*}
where the upper-level optimization problem is defined by upper-level objective $F(\cdot)$ and upper-level constraints encoded in $G(\cdot)$ and $H(\cdot)$. The lower-level optimization problem, defined by lower-level objective $f_0$, with lower-level constraints $\{f_i(\cdot)\}_{i=1}^{m}$ and $\{h_i(\cdot)\}_{i=1}^{p}$, is embedded as a constraint in the upper-level optimization problem. The upper-level and lower-level decision variables are $x_u$ and $x_l$, respectively.
In Fig.~\ref{fig:bo} we illustrate a simple bilevel optimization problem where $x_u\equiv y$ encodes the constraints and $x_l$ is the lower-level variable to be solved for every instance of $y$.
In this problem for every $y$ the constraint is different, and so is the optimal solution.
In general, both the cost function and constraint may depend on the upper level variable $x_u$ but here only the constraint depends on $x_u$.
The goal in this problem is to find the optimal $y$ such that the corresponding lower-level problem has the smallest cost.
Our analytic gradient computes the gradient of the lower level optimal cost with respect to the upper level variable $x_u$.
\end{definition}

Following this definition, we rewrite \eqref{problem:bezier_optimization} as:

\begin{equation}
    \begin{array}{lll}
    \underset{c, y}{\mbox{minimize}}   & J(c,y) = c^T P(y) c + w \mathds{1}^T y \\
    \mbox{subject to} & c \in \underset{c}{\argmin} \{ J(c,y): G(y) c \leq h, \; H(y) c = m \} \\
    & A y \leq b \\
    & C y = d.
    \end{array}
    \tag{TOBC-BO}
    \label{problem:bilevel_optimization}
\end{equation}

Note that although the objective functions $J(c,y)$ remain the same in both the lower-level (also the first constraint) and upper-level optimization problem, $y$ is fixed in the lower-level optimization problem but becomes the optimization variable in the upper-level optimization problem.  The lower-level problem is also a quadratic program (QP) because $J(c,y)$ is quadratic in $c$ when $y$ is fixed.

Our solution strategy is to use constrained gradient descent on the function $J^\star(y) = J(c^\star(y),y)$ with $c^\star$ minimizing the QP for every time allocation $y$, i.e.,
\begin{equation}
 c^\star(y) \in \underset{c}{\argmin} \{ J(c,y): G(y) c \leq h, \; H(y) c = m\}.
 \label{eq:lower_level_problem}
\end{equation}
Gradient descent has, indeed, been used to solve bilevel optimization problems~\cite{sinha2018review}, and our framework is a variant of this method. Note that in Eq.~\eqref{problem:bilevel_optimization} and \eqref{eq:lower_level_problem} we express $c\in \argmin $ since the solution to the lower problem is in general a set. 
However, we prove in Sec.~\ref{sec:convergence_analysis} that the optimal solution to TOBC is unique.
Given a feasible $y \in \mathbb{R}^n$, we find a direction $-\nabla_y J^\star(y) \in \mathbb{R}^{n}$ and a step length $\alpha$ that can make a sufficient decrease in $J^\star (y)$ while maintaining the feasibility of the new point $y_\mathrm{new} = y - \alpha \nabla_y J^\star(y)$.  The key issue with this approach is obtaining the gradients, which we address in the next section. 

\Gao{Because gradient descent does require a feasible initial guess of $y$, it is worth describing how such a feasible point can be determined. For TOBC, feasibility requires meeting velocity and acceleration limits, which requires enough time to be allocated to segments. 
If the initial and final states are static (with zero velocity and acceleration), one can increase the total trajectory time by scaling the time allocation to make it feasible. In other cases, however, an initial guess may not be feasible. One possible approach to mitigate this problem is to define a relaxed inner problem with slack variables if the initial guess is infeasible. We leave this problem to be addressed in future work.}

\subsection{Gradient Computation}
We use a key result from sensitivity analysis of parametric nonlinear programming (NLP)~\cite[Thm 2.3.3]{Jittorntrum1978} to derive the gradient $\nabla J^\star (y)$. In our problem, the lower-level objective is the same as the upper-level one, which allows us to derive gradients of the upper-level decision variables through sensitivity analysis.
For brevity, we give results of first-order sensitivity analysis and refer readers to \cite{Jittorntrum1978} for details.

\begin{theorem}
Consider the problem of finding the local solution $c(y)$ of a parametric NLP problem:
\begin{equation*}
    \begin{array}{ll}
    \underset{c}{\mbox{minimize}}   & J(c, y) \\
    \mbox{subject to} & g_i(c, y) \leq 0, \quad i = 1, \ldots, m \\
    & h_j(c, y) = 0, \quad j=1, \ldots, p
    \end{array}
\end{equation*}
where $c$ is the vector of decision variables and $y \in \mathbb{R}^n$ is a parameter vector.
Let $c^\star(y_0)$ be a locally optimal solution, and let  $\lambda$ and $\nu$ be Lagrange multipliers associated with $g(\cdot)$ and $h(\cdot)$, respectively. 

If the following conditions hold:
\begin{enumerate}
    \item functions $J(\cdot)$, $g_i(\cdot)$ (for all $i$) and $h_j(\cdot)$ (for all $j$) are twice continuously differentiable in $c$, and their gradients w.r.t. $c$ and the constraints $g_i(\cdot)$ (for all $i$) and $h_j(\cdot)$ (for all $j$) are once continuously differentiable in $y$ in a neighborhood of $(c^\star, y_0)$,
    \item objective $J(c,y)$ is twice continuously differentiable in $(c, y)$ near $(c^\star, y_0)$,
    \item the strong second-order sufficient conditions (SSOSC) hold at $c^\star(y_0)$,
    \item the gradients $\nabla g_i(c^\star, y_0)$ (for $i$ such that $g_i(c^\star, y_0) = 0$) and $\nabla h_j(c^\star, y_0)$ (for all $j$) are linearly independent,

\end{enumerate}

then in a neighborhood of $y=y_0$, the gradient of the objective is
        \begin{equation}
        \nabla_{y} J^\star(y) = \nabla_{y}J + \sum_{i=1}^{m} \lambda_i(y) \nabla_{y}g_i + \sum_{j=1}^{p} \nu_j(y) \nabla_{y}h_j.
        \label{eq:gradient_computation}
        \end{equation}
\label{theorem:gradient}
\end{theorem}

Moreover, similar theorem in \cite{fiacco1983introduction} further requires the Strict Complementary Slackness (SCS), i.e., $\lambda_i > 0$ when $g_i(c^\star, y_0) = 0$ but only requires second-order sufficient condition. 
With SCS, the active set which is the collection of inequality constraints where equality holds i.e. $\{i\in\{1,\dots,m\}|g_i(c^*, y_0)=0\}$ does not change. 
Theorem in \cite{Jittorntrum1978} does not require SCS and this means $J^*(y)$ is differentiable even if some constraints switches between being active and non-active.

In our problem, conditions 1) and 2) are satisfied by construction; SSOSC can be proved (see Appendix \ref{app:justification} for details). 
However, condition 4), also known as the Linear Independence Constraint Qualification (LICQ) may fail in some cases. 
Moreover, this condition can only be verified after the NLP is solved and the pattern of active constraints is known.
When LICQ holds, the Lagrangian multipliers are unique and Eq.~\eqref{eq:gradient_computation} computes the exact gradient of the objective.
When LICQ does not hold, however, the Lagrangian multipliers associated with linearly dependent constraints are not unique. 
The Karush-Kuhn-Tucker (KKT) condition is satisfied for infinite number of multipliers.
As a result, the gradient computed by Eq.~\eqref{eq:gradient_computation} may not be unique.
In Sec.~\ref{app:licq_fail}, we show that under the assumption of Slater's condition \cite{boyd2004convex}, Eq.~\eqref{eq:gradient_computation} computes a subdifferential.
Slater's condition only requires the existence of a feasible solution in the interior of the convex feasible set and is not a strict assumption.

\subsection{Solving Bilevel Optimization}

Our algorithm, given in Alg.~\ref{alg:rt}, uses \Mod{ a combination of gradient descent and subgradient descent} to solve \eqref{problem:bilevel_optimization}. It takes an initial guess of the time allocation $y_0$ as input. It then iteratively descends $J^\star (y)$ until some optimality conditions are satisfied or the maximum number of iterations is reached.
\KH{The usual step is a gradient descent step, but if it is detected that the function is nondifferentiable, the algorithm switches to take a subgradient step.  Subgradient descent is widely used in training deep neural networks for non-convex and non-smooth objective functions. Subgradient descent alone, however, is quite slow due to diminishing step sizes.  In our problem, the function is smooth in most regions so  gradient descent with line search is usually far more efficient.  As we shall see in the experiments, the subgradient step is rarely taken, but can kick the algorithm out of states where gradient descent gets stuck. }

\begin{algorithm}
\caption{Refine-Time ($y_0, \alpha_{\rm sub}$)}\label{alg:rt}
\begin{algorithmic}[1]
\State $y \gets y_0, n_{\rm sub}\gets 0, J_{\rm opt}\gets\infty, y_{\rm opt}\gets 0$

\For{$i \gets 0$ to \textit{max-iterations}}
\State $J, \lambda, \nu \gets \text{Solve-QP} (P(y), G(y), h, H(y), m)$ \label{alg:rt:solve_qp}
    \State $g \gets \text{Get-Gradient} (\lambda, \nu)$ \Comment{From Eq. \eqref{eq:gradient_computation}} \label{alg:rt:get_gradient}
    \State $p \gets \text{Project-Gradient} (g, A, b, C, d)$  \label{alg:rt:project_gradient}
    
    \State $\alpha, J, y \gets \text{Line-Search} (y, p)$\label{alg:rt:line_search}
    \If {$\alpha$ not found}
        \State $y\gets y-\alpha_{\rm sub} p / (n_{\rm sub} + 1),  n_{\rm sub}\gets n_{\rm sub} + 1$ \label{alg:rt:subgrad_step}
        \State $J \gets J(y)$
    \ElsIf {optimality-conditions-satisfied}
        \State \textbf{break}
    \EndIf
    \If{$J < J_{\rm opt}$}
        \State $(J_{\rm opt}, y_{\rm opt})\gets (J, y)$ \label{alg:rt:keepbest}
    \EndIf
    
\EndFor
\State \Return $y_{\rm opt}$
\end{algorithmic}
\end{algorithm}

Line \ref{alg:rt:solve_qp} solves a QP problem with a time allocation $y$ and then returns the objective value $J$ and the dual solution (Lagrange multipliers) $\lambda$ and $\nu$. Line \ref{alg:rt:get_gradient} computes the estimated gradient of the objective w.r.t. time allocation $y$ with the Lagrange multipliers $\lambda$ and $\nu$ using Eq.\eqref{eq:gradient_computation}. Line \ref{alg:rt:project_gradient} finds a normalized descent direction from the gradient by projecting the gradient onto the null space of equality constraints  $Cy=d$~\cite{nocedal2006numerical}. 
In the Hard-Time variant that we consider, the constraint is that total time is a constant, and hence the projected gradient is computed as $p=g-\frac{1}{n}\sum{i=1}^n g_i$.

Line \ref{alg:rt:line_search} calls the line search Alg. \ref{alg:ls} to find a suitable step length $\alpha$ that meets the inequality constraints on $y$ and gives sufficient decrease in the objective function.  
If $\alpha$ cannot be found, we assume $y$ is near a non-smooth region, so Alg.~1 takes a subgradient step without checking for sufficient decrease.  If taken, the first subgradient step size $\alpha_{\rm sub}$ is initially set to the initial $\alpha$ at the step when line search fails. 
If a line search step can be found, the following optimality conditions are checked in Line 7:
\begin{enumerate}
    \item Norm of the projected gradient is less than \num{1e-3}. 
    \item The change of the absolute or relative objective function is less than \num{1e-3}.
\end{enumerate}
Since the subgradient step may actually increase the objective function's value, the best solution during all iterations is returned (Lines 12--14).

\begin{algorithm}
\caption{Line-Search ($y_s$, $p$)}\label{alg:ls}
\begin{algorithmic}[1]
\State \textbf{static variable} $\alpha_0$ \label{alg:ls:static_var}
\State \textbf{constant variables} $\tau_g > 1, 1 > \tau_s > 0$ \label{alg:ls:const_var}

\State $\alpha \gets \alpha_0$ \label{alg:ls:initial_alpha}

\For{$i \gets 0$ to \textit{max-iterations}}
    \State $y = y_s - \alpha p$ \label{alg:ls:update_t}
    \State $J, \lambda, \nu \gets \text{Solve-QP} (P(y), G(y), h, H(y), m)$ \label{alg:ls:solve_qp}
    \If {sufficient-decrease-achieved} 
        \If{$i = 0$} \label{alg:ls:start_adaptive}
            \State $\alpha_0 \gets \tau_{g} \alpha$ \label{alg:ls:grow_alpha0}
        \Else
            \State $\alpha_0 \gets \tau_{s} \alpha$ \label{alg:ls:shrink_alpha}
        \EndIf \label{alg:ls:end_adaptive}
        \State \Return $\alpha, J, y$ \label{alg:ls:alpha_found}
    \EndIf
    \State $\alpha \gets \tau_{s} \alpha $
\EndFor
\State \Return \text{``not found''}, $J, y$ \label{alg:ls:alpha_not_found}
\end{algorithmic}
\end{algorithm}
We use an  adaptive backtracking method to find a step $\alpha$ to achieve the Armijo sufficient decrease condition~\cite{nocedal2006numerical}. If it fails to find a decrease, ``$\alpha$ not found'' will be returned as in Line \ref{alg:ls:alpha_not_found}. 
The initial step length $\alpha_0$ defined in Line \ref{alg:ls:static_var} will be updated adaptively. The update strategy is similar to the update of trust region radius in a trust region algorithm \cite{nocedal2006numerical}: $\alpha_0$ will grow or shrink based on the decrease achieved in the first iteration, shown in Line \ref{alg:ls:grow_alpha0} and Line \ref{alg:ls:shrink_alpha}, respectively.
We find adaptive line search useful since it reduces the number of QPs that are solved during line search and time for solving QPs dominates our algorithm's time complexity.

\Gao{
For a candidate step $y \rightarrow y-\alpha p$ where $\alpha$ is the step size and $p$ is the gradient, a non-smoothness occurs when some point along this line segment fails to meet the LICQ condition and first-order Taylor expansion does not well approximate the function value. 
However, we do not attempt to detect every point at which the objective is non-smooth, because the gradient can still make adequate progress (as determined by the sufficient decrease condition) and it is practically computationally expensive to do.  Instead, we decide to trigger the subgradient step only when backtracking line search fails to find a sufficient cost decrease.
The subgradient steps use a standard diminishing step size to guarantee convergence. 
We justify this decision further with the convergence analysis of Sec.~\ref{sec:convergence_analysis}. Moreover, our experiments in Sec.~\ref{sec:subgradient_experiments} suggest that introducing subgradient steps increases the number of iterations, but with the benefit of providing more opportunities to improve the objective.}


\section{Convergence Analysis}
\label{sec:convergence_analysis}
Although gradient descent with line search is convergent in smooth regions~\cite{nocedal2006numerical}, we must study the behavior of Alg.~\ref{alg:rt} in non-smooth regions when it switches to subgradient methods.
The proof is mainly based on results from Davis et al.~\cite{davis2018stochastic} which prove the convergence of the stochastic subgradient descent method for a wide variety of functions including those represented by deep neural networks.
We prove convergence by showing the cost function of the upper optimization in our problem satisfies the conditions of the theorem in \cite{davis2018stochastic}.
We note that the proof relies on the very structure of our problem and is not necessarily true for other bilevel optimization problems. 
Specifically, the main property we use is the convexity of the lower level problem. 

\subsection{Clarke Subdifferential}
\label{app:Subdifferential}

For a locally Lipschitz continuous function $f:\mathbb{R}^d\to \mathbb{R}$, the \emph{Clarke subdifferential} \cite[Ch2, Theorem 8.1]{clarke2008nonsmooth} of $f$ at any point $x$ is the set
\begin{equation}
\partial  f(x)\equiv \text{conv}\left\{ \lim_{i\to\infty}\nabla f(x_i):x_i \xrightarrow{\Omega} x \right\}
\label{eq:clarke_def}
\end{equation}
where $\Omega$ is any full-measure subset of $\mathbb{R}^d$ such that $f$ is differentiable at each of its points; conv means the convex hull of the limit of gradient for all sequences in $\Omega$ approaching $x$
and a point is \emph{(Clarke) critical} if $0\in \partial f(x)$.
An arc $x(t): \mathbb{R}_+\to\mathbb{R}^d$ is called a \emph{trajectory} if it satisfies the differential inclusion
\begin{equation}
\dot{x}(t)\in -\partial f(x(t))\quad \text{for a.e. } t\ge 0
\end{equation}
An iteration sequence is used to track the trajectory
\begin{equation}
x_{k+1}=x_k+\alpha_k(g_k+\xi_k)
\label{eq:subdiff_iteration}
\end{equation}
where $\alpha_k>0$ is a sequence of step sizes that is square summable but not summable, $g_k \in -\partial f(x)$, and $\xi_k$ is noise which is zero in our case. 
Then by \cite[Theorem 3.1]{davis2018stochastic} the sequence generated by Eq.~\eqref{eq:subdiff_iteration} approximates the trajectory of the differential inclusion.

\subsection{Convergence Result}
In order for the trajectory of differential inclusion to converge to a critical point, a sufficient condition is that $f(x)$ must locally Lipschitz and semianalytic \cite[Theorem 5.9]{davis2018stochastic}. 
This condition requires $f(x)$ to be piecewise analytic which fits naturally in our case where different combinations of active inequality constraints result in piecewise functions.
It suffices to show within each piece, that the optimal cost of the lower optimization is an analytic function. 
Considering that the lower problem is a QP which can be solved by inverting a matrix whose entries are analytic functions of the time allocation $y$ given $y>0$, the optimal solution is an analytic function, and, consequently, so is the cost function of lower optimization.

It remains to show that each piece is connected, i.e. the optimal lower cost is continuous with respect to the time allocation $y$.
This can be shown by Theorem 2.1 of \cite{fiacco1983introduction} which states that the optimal cost is continuous as long as: the set-valued mapping from time allocation to the feasible set (of lower optimization) is continuous; the feasible set is compact; and the objective function is continuous. 
These conditions are easy to verify in our problem.
As a result, the subgradient methods being used in Alg.~\ref{alg:rt} converges to a critical point as long as 1) the gradient from Eq.~\eqref{eq:gradient_computation} is indeed a subdifferential even when LICQ fails to hold and 2) the constraints in the upper level optimization do not affect the convergence of the algorithm.
These two conditions are discussed in the next two sections.

\subsection{LICQ Failure}
\label{app:licq_fail}
LICQ fails when, at the optimum point, the linearized equality and active inequality constraints are linearly dependent.
In that case, the Lagrangian multipliers associated with those constraints are non-unique.
In our problem, the lower optimization is QP and, thus, if LICQ fails to hold, it means that some rows of $H$ and $G_{\mathcal{A}}$ are linearly dependent where $G_{\mathcal{A}}$ is the collection of inequality constraints that are active. 
Commercial QP solvers like Sqopt and Mosek have a pre-solve process that eliminates redundant constraints so the actual QP being solved returns unique multipliers and the redundant constraints have multipliers of zero.

When LICQ fails, we have to assume Slater's condition \cite{boyd2004convex} holds. 
The Slater's condition imposes restrictions on the feasible set and requires the existence of a point in the feasible set such that inequality constraints are strictly smaller than 0.
The Slater's condition is not strict and only fails to hold for pathological cases in our problem. One pathological example is due to velocity limit and time limit, the drone has to be at the maximum velocity to travel from one end to another. Considering a 1D single corridor case starting from 0 to 1 within 1 seconds. The velocity limit of 1 m/s requires all velocity inequalities constraints to be active and the feasible set has no interior. This case, however, is close to infeasibility so it rarely occurs. Besides, this case is unstable and small perturbation to $y$ solves the issue.
For convex problems, Mangasarian-Fromovitz Constraint Qualification (MFCQ) \cite{still2018lectures} holds if Slater's condition holds.
Then the optimal cost of the lower problem is directional differentiable \cite[Theorem 7.3]{still2018lectures} in any direction and the directional derivative is given by
\begin{equation}
    \nabla_y J^*(\bar{y}, z) = \min_{\bar{c}\in S(\bar{y})} \max_{\mu \in M(\bar{c}, \bar{y})} \nabla_y L(\bar{c}, \bar{y}, \mu)z
\end{equation}
where $z$ with $\|z\|=1$ is any direction; $S(\bar{y})$ is the set of minimizers which is singleton in our case due to uniqueness of the optimal solution; $\mu$ is the Lagrangian multipliers vector and $M(\bar{c}, \bar{y})$ is the set of valid multipliers that satisfy KKT conditions.
In our problem, the optimum is unique due to SSOSC so we can remove the minimum operator. 
The directional derivative is obtained by essentially solving a linear program to compute the multipliers.
According to the proof within \cite[Theorem 7.3]{still2018lectures}, $M$ is compact if MFCQ holds.
Since constraints are linear, $M$ is essentially a polyhedron defined by equality and inequality constraints of KKT equations, so is the set of values of $\nabla_y L(\bar{c}, \bar{y}, \mu)$ due to its linear dependency on $\mu$.
For all the vertices, there exists a $z$ such that the maximum is obtained at the vertex due to convexity.
In that direction, $\nabla_y L(\bar{c}, \bar{y}, \mu)$ computes the exact gradient.
So according to the definition in Eq.~\eqref{eq:clarke_def}, $\nabla_y L(\bar{c}, \bar{y}, \mu)$ is within the subdifferential.
So the polyhedron of $\{\nabla_y L(\bar{c}, \bar{y}, \mu)|\mu \in M(\bar{c}, \bar{y})\}$ is a subset of the Clarke differential, and so is the gradient computed by any valid $\mu$.

In fact, $\nabla_y L(\bar{c}, \bar{y}, \mu)$ may be unique for all $\mu\in M(\bar{c}, \bar{y})$ such as when the non-dependent constraints do not depend on $y$.
In that case, the exact gradient is computed.
One example in our problem is when two corridors share a face (denoted as the plane $x=b$) and the constraints include $x_i \le b, x_j \ge b, x_i = x_j$ where $x_i$ and $x_j$ is the last and first control points of the two corridors, respectively.
The first two constraints are the control points that are within their corresponding corridors and the third one is path continuity.
At optimum these 3 constraints are always linearly dependent but do not affect the gradient computation since they do not depend on the time allocation.
Moreover, one redundant inequality constraint is removed during the pre-solve process of the optimizers.

\subsection{Constraints in the Outer Optimization}
\label{app:projection_unconstrained}
This section shows that even with constraints in the outer optimization, the theory in Sec.~\ref{app:Subdifferential} can be applied.  The outer optimization in the Soft-Time variant as in Eq.~\eqref{problem:bilevel_optimization:time_penalty} has the form
\begin{equation}
    \begin{array}{lll}
    \underset{y}{\mbox{minimize}}   & J(y) = c^\star + w \mathds{1}^T y \\
    & y \geq 0.
    \end{array}
    \label{problem:unconstrained_bilevel}
\end{equation}
The problem is essentially unconstrained at optimum. The reason is that as $y_i \rightarrow 0$, the cost function approaches $\infty$ and non-positive durations can be rejected during line search.
So the constraint $y \geq 0$ can be ignored in analysis.

The Hard-Time variant as in Eq.~\eqref{problem:bilevel_optimization:min_jerk} has the form
\begin{equation}
    \begin{array}{lll}
    \underset{y}{\mbox{minimize}}   & J(y) = c^\star(y) \\
    & y \geq 0\\
    &\mathds{1}^T y = T_c
    \end{array}
    \label{problem:constrained_bilevel}
\end{equation}
with fixed total traversal time $T_c$.

Here we use a projected (sub)gradient method that computes the (sub)gradient $g\in \mathbb{R}^n$ and projects it onto the hyperplane $\mathds{1}^T g = 0$ so an update
does not change the total traversal time.  The projection operator is given by \[
\mathcal{P} g=g-(\mathds{1}^T g / n)\mathds{1}=(I-\frac{1}{n}\mathds{1}\mathds{1}^T)g
\] 
and the update rule is
\begin{equation}
    y^\prime \gets y-\alpha \mathcal{P} g = y-\alpha (I-\frac{1}{n}\mathds{1}\mathds{1}^T)g.
    \label{eq:projected_update}
\end{equation}

We show that this is equivalent to (sub)gradient descent on an unconstrained problem:
\begin{equation}
    \begin{array}{lll}
    \underset{s}{\mbox{minimize}}   & \tilde{J}(s) = c^\star(y_0 + As) \\
    & y_0 + As \geq 0
    \end{array}
    \label{problem:unconstrained_equivalent_bilevel}
\end{equation}
where $s\in \mathbb{R}^{n-1}$; $y_0$ is the initial time allocation which satisfies $\mathds{1}^T y_0 = T_c$; and $A\in\mathbb{R}^{n\times(n-1)}$ is a matrix forming the orthonormal basis of the null space of $\mathds{1}$.  $A$ can be obtained by the singular value decomposition $\mathds{1} = USV^T$, with $U=[1]$, $S = [\sqrt{n},0,\ldots,0]$, and $V \in \mathbb{R}^{n\times n}$ is an orthogonal matrix with first column $\mathds{1}/\sqrt{n}$ and the remaining rows equal to$A$, that is $V=\left[\begin{array}{c|c}\mathds{1}/\sqrt{n} & A \end{array}\right]$.

The value $y$ corresponding to an iterate $s$ is given by $y_0 + As$. Using the same argument as in the Soft-Time variant, we can show the inequality constraint does not affect convergence. Each iterate of gradient descent satisfies the constraint $\mathds{1}^T (y_0 + As) = T_c$.  Using chain rule, the (sub)gradient of $\tilde{J}$ w.r.t. $s$ is $A^T g$ where $g$ is a (sub)gradient of $J$ at $y_0+As$.
The update rule is thus
\[
s^\prime \leftarrow s - \alpha A^T g 
\]
so the equivalent time allocation update is
\begin{equation}
y' \leftarrow y_0+As'=y_0+As- \alpha A A^T g=y-\alpha AA^T g.
\label{eq:unconstrained_update}
\end{equation}
Using the orthogonality of $V$ we have
\[
I=VV^T=\begin{bmatrix}\mathds{1}/\sqrt{n} & A\end{bmatrix} \begin{bmatrix}\mathds{1}^T/\sqrt{n} \\ A^T\end{bmatrix}=(\mathds{1}\mathds{1}^T)/n+AA^T
\]
which demonstrates that $AA^T = I - \frac{1}{n}\mathds{1}\mathds{1}^T$. Hence, the unconstrained update rule \eqref{eq:unconstrained_update}  is equivalent to~\eqref{eq:projected_update}.

\section{Experiments}

Our algorithm, which is released as an open-source package\footnote{https://github.com/OxDuke/Bilevel-Planner}, is implemented in Python. The QP solvers are called with interfaces to C++ libraries. Since solving QP dominates the running time, our reported computation time is similar to a pure C++ implementation with some overhead from Python.

\subsection{Numerical Experiments}
\label{sec:num_exp}
\begin{figure}[thpb]
    \centering
    \includegraphics[width=0.35\textwidth, angle=90]{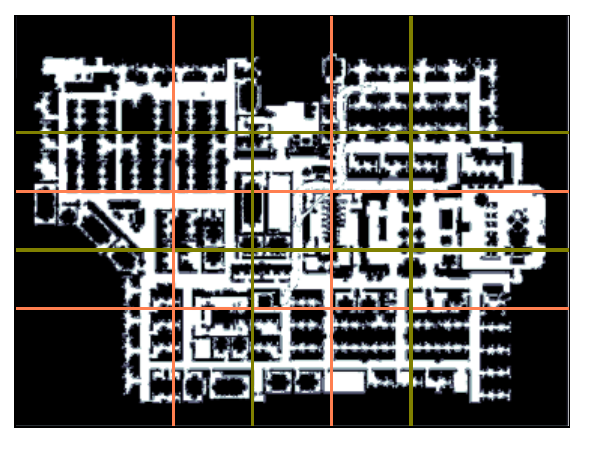}
    \caption{The 2d floor plan used to generate random test problems. Here the orange and light green lines shows where obstacles along $z$ axis are placed.}
    \label{fig:floor_plan}
\end{figure}

\begin{figure}[thpb]
    \centering
    \includegraphics[width=0.33\textwidth, angle=90]{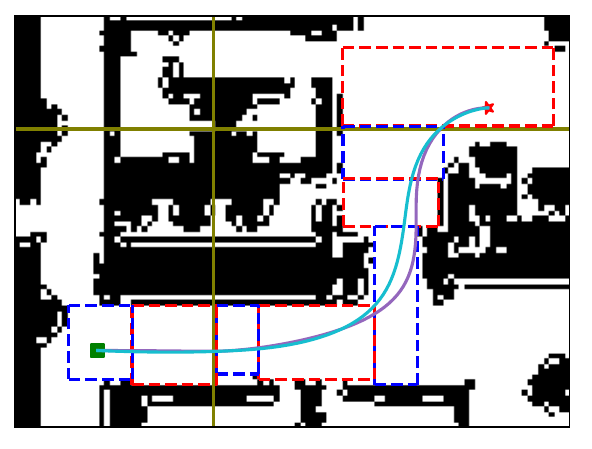}
    \caption{A random problem instance. The start (square), goal (star), and generated corridor (dashed boxes) are shown. The purple and cyan curve are the optimal trajectories with the initial and final time allocations.}
    \label{fig:floor_plan_traj}
\end{figure}

We evaluate our method on random instances of ``indoor flight''. 
We take the indoor building environment from \cite{faust2018prm}, extrude it along $z$ axis, and discretize the $z$ direction into 5 cells.
We then select some rows and columns in the image as shown in Fig.~\ref{fig:floor_plan} and fill the bottom 3 or top 3 cells (along $z$ axis) as obstacles according to the color.
The start and goal are randomly sampled in the free space so in most cases there is movement along $z$ axis.
The environment is shown in Fig.~\ref{fig:floor_plan}.
The complexity of the environment allows us to generate problems with more than 40 segments.
200 problems are randomly generated.
One example is shown in Fig.~\ref{fig:floor_plan_traj}.
\Gao{We note that despite we are exclusively using axis-aligned boxes as safe corridors for simplicity, convex polyhedrons may yield less conservative results and our method is able to handle it as well. But there is a trade-off between the complexity of corridors and optimality of the solution.}

We solve the Hard Time variant as in Eq.~\eqref{problem:bilevel_optimization:min_jerk}, with the initial guess of time allocation and fixed total traversal time computed from the heuristic introduced by Gao et al~\cite{gao2018online}. 
In this section, the QP solver is Gurobi \footnote{\url{https://www.gurobi.com/}}, a commercial interior-point QP solver. 
We set a major iteration limit of 50 to limit the total computation time.
The velocity limit is set as 2 m/s.
To handle cases with infeasible initial time allocation, we simply multiply the time allocation by a scalar until feasibility is obtained.
As a benchmark, we compare our analytic gradient with a finite-difference approximation method under the same bilevel optimization framework.
Another method in comparison is to optimize the control points and time allocation simultaneously using nonlinear optimizer SNOPT \cite{snopt77}.
Finally, we compare with direct collocation method \cite{betts1998survey} which discretize the trajectory, formulate an NLP and solve it using SNOPT.
These experiments are carried out on a workstation with a 3.30 GHz Intel Xeon W-2155 processor, using only one thread.

\subsubsection{Finite Difference vs Analytic Gradient}

In this section, we solve the same problem set as before and compare the analytic gradient from Lagrangian multipliers defined in this paper with finite difference approximation, where one gradient computation requires additional $n$ QPs being solved.
The average performance is shown in Tab.~\ref{tab:cmp_fd}.
Clearly, analytical gradients are not only much faster, but also help the optimizer converge to a better solution because of their higher precision. 

\begin{table}[thpb]
    \centering
    \caption{The Mean Total Computation Time, Average Major Iteration Time and Normalized Cost (final cost over the cost from heuristic assignment) of Finite Difference (FD) w.r.t. Analytic Gradient (AG)}
    \label{tab:cmp_fd}
    \begin{tabular}{c|c|c|c}
    \hline
       & Total Time (s) & Avg. Iter. Time (s) & Normalized Cost \\
       \hline
       FD &15.403 & 0.497 & 0.109  \\
       AG & 0.874 & 0.024 & 0.068 \\
       \hline
    \end{tabular}
\end{table}

\subsubsection{Comparison with Joint Optimization}
Joint optimization directly solves Eq.~\eqref{problem:bilevel_optimization:min_jerk} as an NLP using SNOPT~\cite{snopt77}, a general nonlinear solver for sparse, large-scale problems. 
On the contrary, bilevel optimization decouples temporal and spatial variables and solves them hierarchically. 
We provide analytic gradients to SNOPT for solver robustness and explore problem sparsity to the best of our ability. 
We initialize SNOPT with the unrefined time allocation and the spline coefficients computed in the first QP solve. All the stopping criteria are set to default except the optimality tolerance is set to \num{1e-3}.

We obseved that joint optimization is susceptible to the strong non-linearity of the problem and only 11 out of 200 problems converged to a feasible solution, as shown in Tab.~\ref{tab:cmp_jodc}.
Due to the low success rate, it's clearly not suitable for the application.
SNOPT tends to terminate prematurely without converging, and often moves to an infeasible point even though it starts from a feasible solution. We believe this is because the joint spatial and temporal NLP is ill-conditioned. The QP objective function exhibits high-order dependence on timing, and some spatial constraints are very sensitive to the high-order spline coefficients. On the other hand, in the bilevel formulation, the ill-conditioned problem is handled by convex solvers, which are known to be more robust. 
We also note that SNOPT has no guarantee on obtaining a feasible solution while our approach can be terminated at any time and return a feasible solution.
Usually nonlinear optimizers require an initial guess close to the optimum values to converge.
With the same initial guess of time allocation, our method is able to make progress towards optimum while SNOPT moves from a feasible initial guess to non-feasible solutions.

\subsubsection{Comparison with Direct Collocation}

We also compared our proposed method with the Direct Collocation method (DC)~\cite{betts1998survey} to solve problem~\eqref{problem:bilevel_optimization:min_jerk} as an alternate NLP formulation.  DC optimizes over discretized states $[p(t), \dot{p}(t), \ddot{p}(t)]$ where $p(t)$ is position and control is $u\equiv \dddot{p}(t)$ at each collocation grid point $t_0,\ldots,t_N$. 
To adjust the timing of each segment, the initial and final times of each segment are introduced as additional decision variables. 
The state and control trajectories are optimized simultaneously with segment times. Each segment has a fixed grid size. See Appendix~\ref{app:DirectCollocation} for the details.

Once again, we use SNOPT to solve the NLP.  The major iteration limit and total iteration limit are set as 500 and 5000, respectively, to keep the total computation time manageable. Due to the different NLP formulation, the final objective value from DC cannot be directly compared to other approaches so we just compare success rate instead. 
Direct collocation does not perform well on this problem and only 44 out of 200 problems converge to an optimal solution, as shown in Tab.~\ref{tab:cmp_jodc}. 
\Gao{
We observed that with a higher iteration limit, DC can achieve a higher success rate. 
However, the average computation time with the current settings is already 7.12\,s, which is unsuitable for responsive quadrotor flight.
}

\begin{table}[thpb]
    \centering
    \caption{Success Rate of Bilevel Optimization, Joint Optimization, and Direct Collocation}
    \label{tab:cmp_jodc}
    \begin{tabular}{c|c|c}
    \hline
        Bilevel Optimization & Joint Optimization  &  Direct Collocation  \\
        \hline
       200/200 & 11/200 & 44/200 \\
       \hline
    \end{tabular}
\end{table}

\subsection{Scalability Study}
For complex environments, the number of safe corridors may be as large as around 50 and thus imposes a challenge to the computational efficiency.
We perform an empirical scalability study to see how our framework performs with increasing number of segments using the same problems in the last section.
We also study the effect of QP solvers and compare two types of QP solvers: active-set solver Sqopt~\cite{sqopt77} and interior-point solver Gurobi.
Both solvers are designed for sparse and large-scale problems.
Our previous test suite \cite{accversion} showed that Sqopt performs better when the average number of segments is below 10. In this paper we want to study how it performs when the problem has more segments. 
We use the same setting as the last section with the two QP solvers.

It turns out the performance of our algorithm in terms of cost function is quite consistent with the two solvers.
Numerical errors result in slight difference in the progress of the algorithm in each solver.
We note that we set an iteration limit of 50 and around 60 problems out of 200 are terminated after reaching the iteration limit.
On average each iteration requires 1.2 QPs being solved.
This also indicates the total computation time is roughly proportional to the average time of each QP solving.
However, the computation time is quite different for the two solvers, indicating different scalability in QP solving time.

The computation time is different since Sqopt and Gurobi have different scalability with the number of segments which is proportional to the number of optimization variables and constraints. 
Gurobi implements an interior-point method so its iteration number is roughly constant.
In each iteration a sparse block-diagonal matrix is factorized which takes time linearly to the number of segments.
Sqopt implements the active-set method which does not scale well since it takes more iterations to identify the correct active set.
However, it is faster than Gurobi when the number of segments is below 10, which is consistent with the results in \cite{accversion}.
If finite-difference is used to estimate the gradient, the algorithm has quadratic and cubic scalability when the QP solver is Gurobi and Sqopt, respectively.

\begin{figure}[thpb]
    \centering
    \includegraphics[width=0.4\textwidth]{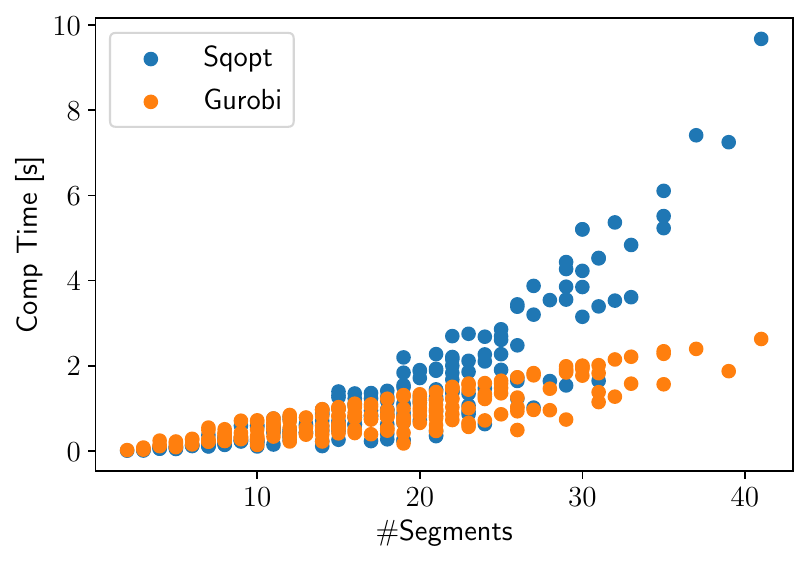}
    \caption{The total computation time of our algorithm as a function of the number of segments in random problems.}
    \label{fig:alg_time_scalability}
\end{figure}

\begin{figure}[thpb]
    \centering
    \includegraphics[width=0.4\textwidth]{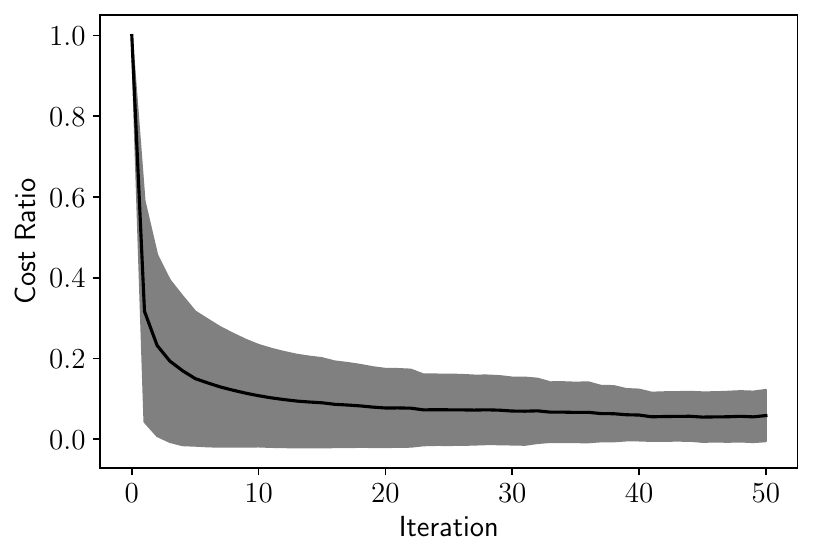}
    \caption{The profile of the average (solid line) and standard deviation (shaded) of the cost ratio over the initial cost. The cost ratio is computed by dividing the cost at current iteration by the initial cost.}
    \label{fig:avg_cost_profile}
\end{figure}

The total computation time scales similarly to average QP times since similar numbers of QPs are solved, as shown in Fig.~\ref{fig:alg_time_scalability}.
The total computation time scales linearly when Gurobi is the QP solver.
The total computational time shows the potential of applying our algorithm in real-time since the major improvement of cost functions occurs in the first few iterations, evidenced in Fig.~\ref{fig:avg_cost_profile}.
For large problems with limited computation time, a smaller iteration limit has to be used, although it does not affect the cost function too much.

\begin{figure}[thpb]
    \centering
    \includegraphics[width=0.4\textwidth]{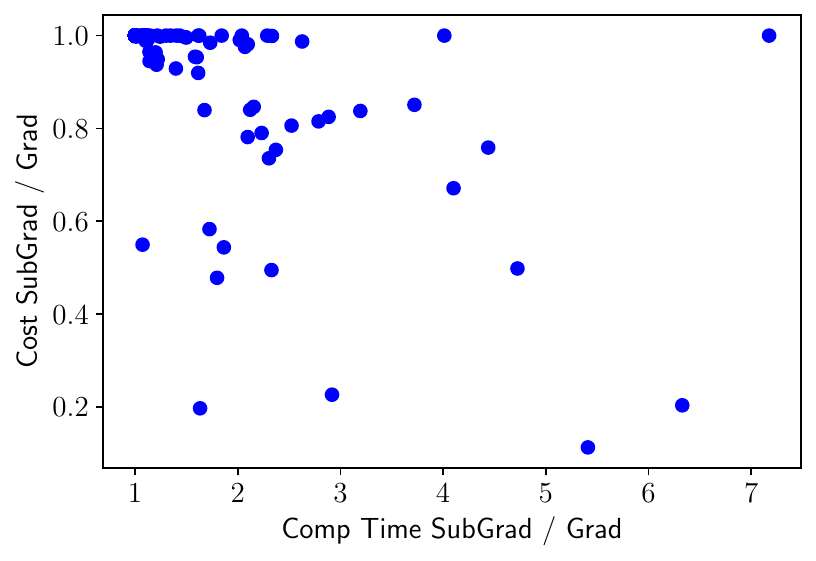}
    \caption{\Gao{Enabling subgradient descent can improve cost when pure gradient descent gets stuck. The $x$ and $y$ axis are the ratios of computation time and cost function, respectively, of Alg.~1 with subgradient descent enabled vs the disabled variant.  Lower is better for both axes.}}
    \label{fig:licq_nolicq_cmp}
\end{figure}

\Gao{
\subsection{Effect of Subgradient Step}
\label{sec:subgradient_experiments}
Although the numerical experiments above are conducted on challenging 3D problems, we found that the subgradient step of Alg.~1 is only triggered once. To better examine the effect of subgradient steps, we designed a problem set that triggers subgradient steps more frequently. These variants project the environments into 2D to eliminate vertical movement, and velocity limits are disabled.  808 random problem instances of this form were generated, and we compared Alg.~1 with and without Lines 9--12.

The results are shown in Fig.~\ref{fig:licq_nolicq_cmp}, with both Sqopt and Gurobi as the underlying QP solver.  A subgradient step was taken in 15 problems using Squopt, and in 65 problems using Gurobi.
The discrepancy between solvers is due to the  slightly different gradient estimation, which may lead to different algorithmic behavior overall.  The plot combines results from subgradient steps taken with either QP solver.  For many problems, the subgradient step decreases the cost substantially.  However, it does pay a price in terms of increased computation time.
The reason is that the gradient-only variant stalls out more quickly, while the subgradient variant continues to optimize and make progress.}

\subsection{Physical Experiments}

We validate our planner on an indoor obstacle avoidance scenario using a commercially available small-scale quadrotor, Crazyflie 2.1\footnote{https://www.bitcraze.io/}. A video that compiles all the physical experiments is provided as supplementary material.

\begin{figure}[thpb]
    \centering
    \includegraphics[width=0.48\textwidth]{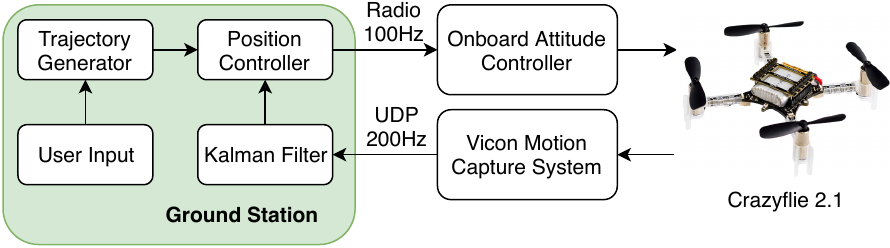}
    \caption{Physical quadrotor system setup}
    \label{fig:system}
\end{figure}

\begin{figure}[thpb]
\centering
\includegraphics[width=0.48\textwidth]{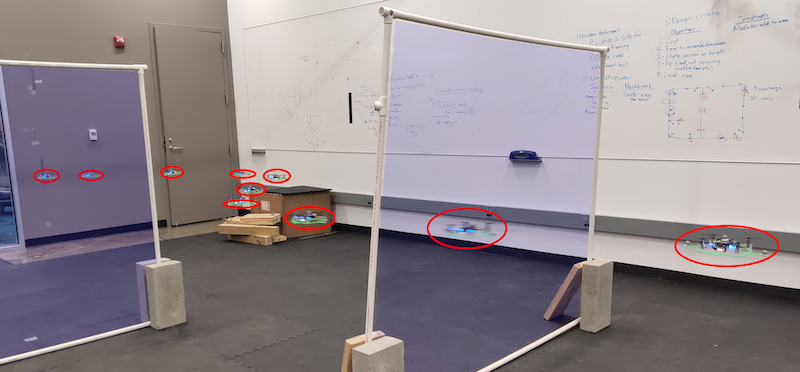}
\caption{\Mod{Obstacle layout for physical quadrotor  experiments, with the space within the frames (tinted blue) are treated as obstacles. Evenly spaced frames from the executed trajectory are overlaid, with the quadrotor circled in red.}  }
\label{fig:physical_experiment}
\end{figure}

The system setup is shown in Fig.~\ref{fig:system}.
The position of the quadrotor is captured by the Vicon\footnote{https://www.vicon.com/} motion capture system and transmitted to ground control station using Ethernet at 200Hz. 
The raw data stream from Vicon goes through a Kalman filter and then serves as feedback for a position controller on the ground control station. 
The position controller is a feedfoward-feedback controller, with the feedforward term computed from the reference trajectory and its time derivative thanks to the differential flatness property, and the feedback term computed from a proportional-integral-differential (PID) controller.

As shown in Fig.~\ref{fig:physical_experiment}, we set up two frames as walls, and considered them as obstacles in our algorithm. We run 3 experiments to demonstrate the effectiveness and real-time capability of our algorithm.
Admittedly, in these physical experiments the number of segments is not large. 
For problems with more segments, a smaller iteration limit can be used without losing too much optimality.
Indeed, the property of any-time feasibility of our algorithm is suitable for challenging problems with many segments since the user can set arbitrary number of iterations.

\subsubsection{Comparison to time allocation heuristics}
Our algorithm plans a faster trajectory while achieving the same control effort (jerk) as Gao el al.~\cite{gao2018online}, which uses time allocation heuristics. The quadrotor starts at some initial position and is asked to travel to a target position chosen by a human operator using an Rviz\footnote{http://wiki.ros.org/rviz} interface while avoiding obstacles. 
Both methods are set up to solve the Hard Time variant \eqref{problem:bilevel_optimization:min_jerk}, but after running our algorithm, we scale the total traversal time $T$ until the jerks of two trajectories becomes the same.  Table.~\ref{tab:point_to_point_jerk} indicates that our algorithm yields a 12\% shorter and 18\% faster trajectory with equivalent jerk.  

\begin{table}[h]
\centering
\renewcommand{\arraystretch}{1.1}
\caption{Comparing against heuristic time assignment}
\label{tab:point_to_point_jerk}
\begin{tabular}{c|c|c|c}
\hline
Method & Length & Traversal Time & Jerk \\ \hline
\textbf{Ours} & \textbf{5.15}\,m & \textbf{4.36}\,s & 39 \\ 
Gao et al. & 5.82\,m & 5.32\,s & 39 \\ \hline
\end{tabular}
\end{table}

\subsubsection{Controlling aggressiveness using time penalty $w$}
Next we show the Soft Time variant \eqref{problem:bilevel_optimization:time_penalty} can handle objectives with various time penalties to control aggressiveness. Results are summarized in Table~\ref{tab:point_to_point_time}.  These indicate that, as expected, when the weight penalty increases, trajectories become faster and more jerky.  Also, computation time is not significantly affected by the weight parameter.

\begin{table}[h]
\centering
\renewcommand{\arraystretch}{1.1}
\caption{Comparing motion aggressiveness parameters}
\label{tab:point_to_point_time}
\begin{tabular}{c|c|c|c}
\hline
Weight ($w$) & Traversal Time & Computation time & Jerk \\ \hline
10 & 5.60\,s & 10.8\,ms & 11.2 \\ 
20 & 4.96\,s & 10.7\,ms & 19.9 \\ 
40 & 4.42\,s & 10.7\,ms & 36.1 \\ 
80 & 4.01\,s & 9.3\,ms & 64.7 \\ \hline
\end{tabular}
\end{table}

\subsubsection{Tracking a dynamic goal}
The final experiments show the algorithm running in real-time, where the quadrotor tracks a dynamic goal moved by a human while avoiding obstacles. The goal is tracked by Vicon and is used for replanning at 3Hz. Trajectories are generated using the Soft Time variant \eqref{problem:bilevel_optimization:time_penalty} with weight $w=80$.  In these experiments, each optimization takes less than 15\,ms.

\section{Conclusion}
We presented a novel bilevel optimization approach to UAV trajectory optimization, which analytically calculates the gradient of the objective function w.r.t. temporal variables. 
The optimization method takes into account the non-smoothness of the upper-level optimization problem.
Our results show that this approach achieves real-time performance and higher quality trajectories than state-of-the-art heuristics.  
Our method can handle both hard time variant with fixed total time and soft time variant where the weight is used to adjust trajectory aggressiveness. 
It can be useful in multiple contexts such as formation flight and tracking dynamic targets.


Future work may include accelerating the gradient descent by exploiting the structure of the problem. For example, acceleration may be achieved through using Newton or Quasi-Newton methods.
We are also interested in studying extensions of the bilevel optimization approach to other robotic applications like autonomous vehicles and legged locomotion.


\appendices
\numberwithin{equation}{section}
\setcounter{equation}{0}

\section{Proof of SSOSC}
\label{app:justification}
We prove the second-order sufficient conditions (SSOSC)  holds in TOBC where jerk is minimized amongst B\'ezier curves of order 6. For brevity we drop dependency on $y$. 

    SSOSC \cite{fiacco1983introduction} states that the Hessian of the Lagrangian evaluated at the optimal point is positive definite on the null space of the gradients of all the active constraints, i.e.,
    \begin{equation*}
        w^T \nabla^2_{cc} L(c^\star, \lambda^\star, \nu^\star) w > 0, \; \forall w \neq 0 \,\text{s.t.}\, G_{\mathcal{A}} w = 0; H w=0
    \end{equation*}
    where $G_{\mathcal{A}} $ collects the rows of active inequality constraints and $L(\cdot)$ is the Lagrangian:
    \begin{equation*}
        L(c, \lambda, \nu) = c^TPc + \lambda^T Gc + \nu^T H c,
    \end{equation*}
    where $\lambda$ and $\nu$ are the associated Lagrange multipliers.
  
    Our proof shows that there does not exist any $w \neq 0$ that lies in the null space of $P$ and $H$ simultaneously, which proves SSOSC because $\nabla^2_{cc} L(c^\star, \lambda^\star, \nu^\star) = P$, and since $P$ is a symmetric positive semi-definite matrix, $w^TPw=0$ implies $Pw=0$. (Note that we ignore $G_{\mathcal{A}}$, so the proof holds regardless of which inequality constraints are active. )
    
    Since each of the  $x, y, z$ dimensions can be decoupled, we show the proof in one dimension without loss of generality.
    Matrix $P$ is block diagonal, with blocks denoted $P_i\in \mathbb{R}^{7\times7}$, whose entries depend on the duration of the corresponding segment. 
    It can be be shown that $$\mathcal{N}_{P_i} = \begin{bmatrix}
        1 & 1 & 1 \\
        1 & 2 & 4 \\
        1 & 3 & 9 \\ 
        1 & 4 & 16 \\
        1 & 5 & 25 \\ 
        1 & 6 & 36 \\
        1 & 7 & 49
        \end{bmatrix}$$
    is a basis for the null space of $P_i$, regardless of segment duration. 
    The physical meaning for each column of $\mathcal{N}_{P_i}$ is not moving at all, moving with constant velocity, and moving with constant acceleration, respectively, since these motions induce zero jerk.
    Then, for a problem with $n$ segments, the matrix
    \[
    \mathcal{N}_P = \begin{bmatrix}[c:c:c]
    \mathcal{N}_{P_i} &  &  \\ 
      & \ddots  &   \\ 
      &  & \mathcal{N}_{P_i}
    \end{bmatrix}
    \]
    is a basis of the null space of $P$, and has rank $3n$.
    
    The linear equality constraints include initial and final position, velocity, and acceleration which accounts for 6 constraints. Continuity of position, velocity, and acceleration at conjunction points provide another $3(n-1)$ constraints for a problem with $n$ segments.
The linear equality constraint matrix $H$ is defined as
\begin{equation}
H = \begin{bmatrix}[ccc:ccc:c:ccc]
L_1 & 0 &  &  &  & & & &  & \\ 
 & 0 & R_1&-L_2 &0  &  & & &  & \\ 
 &  &  &  &0 & R_2&  & &  \\ 
 &  & & & &  &  \ddots&  && \\ 
 & && &  &  &  &  -L_n & 0& \\
 & && &  &  &  &   & 0& R_n
\end{bmatrix}.
\label{eq:HMat}
\end{equation}
with $3\times 3$ blocks:
\begin{equation*}
\begin{aligned}
L_i&=\begin{bmatrix}
1 &  & \\ 
 -d/\Delta t_i& d/\Delta t_i & \\ 
d(d-1)/\Delta t_i^2 & -2d(d-1)/\Delta t_i^2  & d(d-1)/\Delta t_i^2
\end{bmatrix}, \\
R_i&=\begin{bmatrix}
0 &  &1 \\ 
 0& -d/\Delta t_i & d/\Delta t_i \\ 
d(d-1)/\Delta t_i^2 & -2d(d-1)/\Delta t_i^2  & d(d-1)/\Delta t_i^2
\end{bmatrix},
\end{aligned}
\end{equation*}
and each $0$ block is of size $3\times (d-5)$. Here, $\Delta t_i$ is the time allocated to the $i$'th segment and $d=6$ is the spline degree.  
    
    Now we show that $H\mathcal{N}_P$ has full column rank. First, split $\mathcal{N}_{P_i}$ into blocks $[Q_1^T, Q_2^T, Q_3^T]^T$ where $Q_1, Q_3 \in \mathbb{R}^{3 \times 3}$ and $Q_2 \in \mathbb{R}^{1 \times 3}$. Note that $Q_1$ and $Q_3$ are square matrices of full rank. With this notation we express the product
    \[
    H\mathcal{N}_P = \begin{bmatrix}[cccc]
    L_1Q_1 &  &  & \\ 
     R_1 Q_3 & -L_2Q_1 &  & \\ 
     & R_2Q_3 &\ddots  &  \\ 
     & &\ddots& -L_nQ_1 \\
     & && R_n Q_3
    \end{bmatrix}.
    \]
    Observe that each block is non-singular because it is a product of two square matrices of full rank. The upper $3n$ by $3n$ part is full column rank since each block is non-singular. Therefore, $H\mathcal{N}_P$ has full column rank.
    
    Consequently, any $w\neq 0$ in the nullspace of $P$ can be expressed as $w= \mathcal{N}_P v$ with $v\neq 0$, which shows that $H w \neq 0$. Hence there does not exist $w \neq 0$ that lies in the null space of $P$ and $H$ simultaneously, which proves SSOSC as desired.

\section{Direct Collocation Implementation}
\label{app:DirectCollocation}
This section provides details of the direct collocation (DC) method implemented for our numerical experiments.  Our implementation follows the standard \emph{Hermite-Simpson} collocation method, see e.g. Betts~\cite[Section 5.1]{betts1998survey}. DC works by discretizing a continuous-time trajectory optimization problem into a NLP with discretized states and controls as decision variables. 
However, direct collocation is rarely used in UAV trajectory optimization since it does not take advantage of the differential flatness and is usually used directly with the system's actual dynamics equations.
In order to use DC to minimize trajectory jerk, we define the system state $x(t) = [p(t), \dot{p}(t), \ddot{p}(t)]^T \in \mathbb{R}^9$ as a stacked  vector of position $p(t) \in \mathbb{R}^3$, velocity $\dot{p}(t) \in \mathbb{R}^3$ and acceleration $\ddot{p}(t) \in \mathbb{R}^3$. The control input $u(t)$ is chosen to be the jerk $u(t) = \dddot{p}(t) \in \mathbb{R}^3$. 

The system dynamics can be written as:
\begin{equation}
    \dot{x}(t) = f\big(x(t), u(t)\big) = Ax(t) + Bu(t)
    \label{eq:dc:continous_dynamics}
\end{equation}
where
\begin{equation*}
    A = 
    \begin{bmatrix}
    0 & I_{3\times 3} & 0 \\
    0 & 0 & I_{3\times 3} \\
    0 & 0 & 0 
    \end{bmatrix},
    B = 
    \begin{bmatrix}
    0 \\ 0 \\ I_{3 \times 3}
    \end{bmatrix}.
\end{equation*}

Consider a safe corridor with $N$ segments, we set up a $N$-phase trajectory optimization problem. For the $i$'th phase, the trajectory is constrained to stay in the $i$'th convex region. Continuity constraints up to acceleration are applied between consecutive segments. This formulation matches with TOBC. The decision variables include:
\begin{enumerate}
    \item Duration of each phase: $\Delta t_i, i = 1, \dots, N$,
    \item Discretized state trajectory of the UAV for each phase: $x_i(t), t \in [0, \Delta t_i], i = 1, \dots, N$,
    \item Discretized control trajectory for each phase: $u_i(t), t \in [0, \Delta t_i],  i = 1, \dots, N$,
\end{enumerate}
and the cost function is
\begin{equation}
    J=\sum_{i=1}^N \int_{0}^{\Delta t_i} \|u_i(\tau)\|^2 d\tau + w \Delta t_i.
\end{equation}
The rest follows the standard DC formulation~\cite{betts1998survey}.

\ifCLASSOPTIONcaptionsoff
  \newpage
\fi

\bibliographystyle{IEEEtran}
\bibliography{main.bib}

\begin{IEEEbiography}[{\includegraphics[width=1in,height=1.25in,clip,keepaspectratio]{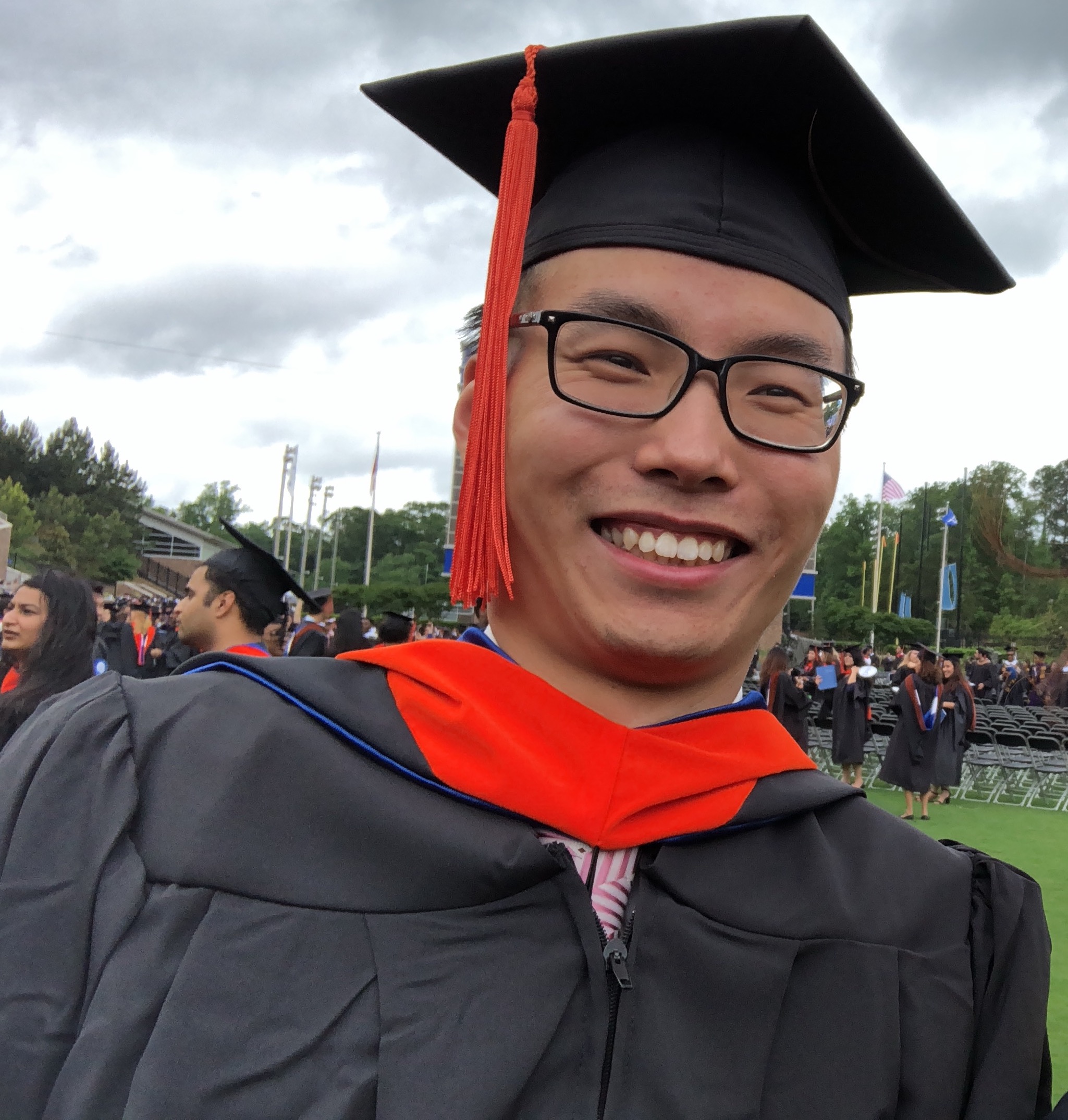}}]{Weidong Sun}
received his master's degree from Duke University, Durham, NC, USA, in 2019, under the sponsorship of the China Scholarship Council. He received his bachelor's degree from Wuhan University of Technology, Wuhan, Hubei, China, in 2017. He is currently a Robotics engineer at XYZ Robotics Inc. at Shanghai, China.
His research focus is on trajectory optimization and motion planning.
\end{IEEEbiography}

\begin{IEEEbiography}[{\includegraphics[width=1in,height=1.25in,clip,keepaspectratio]{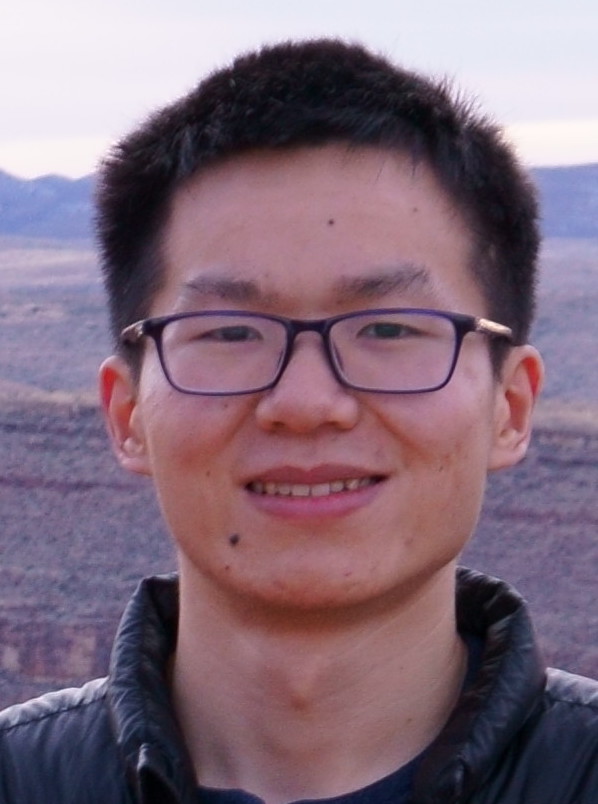}}]{Gao Tang}
is PhD student at University of Illinois at Urbana-Champaign in the Department of Computer Science starting from 2019. 
He was a PhD student at Duke University from 2016 to 2019.
He received his bachelor's and master's degree in Aerospace Engineering from Tsinghua University in 2014 and 2016, respectively. 
His research is mainly focused on motion planning and trajectory optimization.
\end{IEEEbiography}

\begin{IEEEbiography}[{\includegraphics[width=1in,height=1.25in,clip,keepaspectratio]{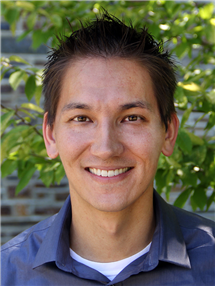}}]{Kris Hauser}
is Associate Professor at University of
Illinois at Urbana-Champaign in the Department of
Computer Science and the Department of Electrical
and Computer Engineering. He received his PhD
in Computer Science from Stanford University in
2008, bachelor’s degrees in Computer Science and
Mathematics from UC Berkeley in 2003, and was
a postdoc at UC Berkeley. He has also held faculty
positions at Indiana University from 2009--2014 and
Duke University from 2014--2019. He is a recipient
of a Stanford Graduate Fellowship, Siebel Scholar Fellowship, Best Paper
Award at IEEE Humanoids 2015, and an NSF CAREER award.
\end{IEEEbiography}

\end{document}